\documentclass[conference]{IEEEtran}
\IEEEoverridecommandlockouts
\usepackage{cite}
\usepackage{amsmath,amssymb,amsfonts}
\usepackage{algorithmic}
\usepackage{graphicx}
\usepackage{textcomp}
\usepackage{xcolor}
\usepackage{subfigure}
\usepackage{caption}
\usepackage{url}

\def\BibTeX{{\rm B\kern-.05em{\sc i\kern-.025em b}\kern-.08em
    T\kern-.1667em\lower.7ex\hbox{E}\kern-.125emX}}
\begin{document}

\title{Gated Res2Net for Multivariate Time Series Analysis\\
{\footnotesize }
\thanks{This research was supported by National Natural Science Foundation of China (No.61827810).}
}

\author{\IEEEauthorblockN{1\textsuperscript{st} Chao Yang}
\IEEEauthorblockA{\textit{School of Information Science and Engineering} \\
\textit{Ocean University of China}\\
Qingdao, China \\
yangchao249@hotmail.com}
\and
\IEEEauthorblockN{2\textsuperscript{nd} Mingxing Jiang}
\IEEEauthorblockA{\textit{School of Information Science and Engineering} \\
\textit{Ocean University of China}\\
Qingdao, China \\
jiangmx@ouc.edu.cn}
\and
\IEEEauthorblockN{3\textsuperscript{rd} Zhongwen Guo}
\IEEEauthorblockA{\textit{School of Information Science and Engineering} \\
\textit{Ocean University of China}\\
Qingdao, China \\
guozhw@ouc.edu.cn}
\and
\IEEEauthorblockN{4\textsuperscript{th} Yuan Liu}
\IEEEauthorblockA{\textit{School of Information Science and Engineering} \\
\textit{Ocean University of China}\\
Qingdao, China \\
liuyuan6787@stu.ouc.edu.cn}

}

\maketitle

\begin{abstract}
Multivariate time series analysis is an important problem in data mining because of its widespread applications. With the increase of time series data available for training, implementing deep neural networks in the field of time series analysis is becoming common. Res2Net, a recently proposed backbone, can further improve the state-of-the-art networks as it improves the multi-scale representation ability through connecting different groups of filters. However, Res2Net ignores the correlations of the feature maps and lacks the control on the information interaction process. To address that problem, in this paper, we propose a backbone convolutional neural network based on the thought of gated mechanism and Res2Net, namely Gated Res2Net (GRes2Net), for multivariate time series analysis. The hierarchical residual-like connections are influenced by gates whose values are calculated based on the original feature maps, the previous output feature maps and the next input feature maps thus considering the correlations between the feature maps more effectively. Through the utilization of gated mechanism, the network can control the process of information sending hence can better capture and utilize the both the temporal information and the correlations between the feature maps. We evaluate the GRes2Net on four multivariate time series datasets including two classification datasets and two forecasting datasets. The results demonstrate that GRes2Net have better performances over the state-of-the-art methods thus indicating the superiority.
\end{abstract}


\section{Introduction}
Multivariate time series data are ubiquitous in many real-world applications. Thus, the analysis of multivariate time series data is meaningful and important\cite{keogh2003need}. Multivariate time series data analysis includes classification and forecasting. The applications of classification can be seen in human activity recognition\cite{anguita2012human}, equipment maintenance planning\cite{liu2018time}, heart disease prediction\cite{masethe2014prediction}, etc., while the applications of forecasting can be seen in air quality forecasting\cite{kolehmainen2001neural}, renewable energy management\cite{doucoure2016time}, power consumption prediction\cite{le2019improving}, etc. The characteristics of multivariate time series data are the natural temporal ordering and the correlations between the variables. Therefore, how to effectively extract and utilize the temporal features and the correlations between variables are essential for multivariate time series analysis.
\\Traditional machine learning algorithms based on statistics theory have been widely used in multivariate time series analysis such as K- Nearest Neighbor (KNN), Dynamic Time Warping (DTW), Support Vector Machine (SVM), etc. For these algorithms, feature engineering is vital for achieving satisfied performance. Because the quality of feature engineering highly depends on the expertise hence resulting in challenges and time expenses.
\\With the advent of big data era, more and more multivariate time series data are accessible for training which leads to the popularity of deep learning models. Deep learning models allow automatic feature extraction and end-to-end learning hence are less dependent on feature engineering which is an advantage compared with traditional machine learning algorithms. Recurrent Neural Network (RNN) is a typical kind of deep learning model and has been widely used in multivariate time series analysis tasks\cite{freeman2018forecasting,che2018recurrent,malhotra2017timenet}. Compared with RNN, Convolutional Neural Network (CNN) is often used in computer vision tasks but less common for time series analysis. CNN can be fully parallelized during training to better exploit the GPU hardware. Thus, recent works have tried to apply CNN in sequence modeling\cite{gehring2017convolutional} and time series classification\cite{zhao2017convolutional}. 
\\The component of a model for feature extraction is called the backbone network whose output is the feature map of the input data. Famous backbone networks includes VGG\cite{simonyan2014very}, Inception\cite{szegedy2015going}, ResNet\cite{he2016deep}, etc. Recent studies have shown that typical CNN backbone networks which are firstly designed for image processing such as ResNet and FCN\cite{long2015fully} are capable for time series analysis\cite{fawaz2019deep}. The neurons in CNN respond to stimuli in a restricted region, often called receptive filed, of the visual field hence CNN is capable for feature extraction in a particular region. Since CNN learns the representations for fixed size contexts, for a better feature extraction ability, it is necessary to design a network architecture to achieves multi-scale feature representations. 
\\Res2Net\cite{gao2019res2net} is a recently developed backbone CNN. It improves the multi-scale representation ability at granular level by utilizing multiple available receptive fields. In a Res2Net module, groups of filters are connected in a hierarchical residual-like style for multi-scale receptive fields which is conducive for global and local feature extraction and the information interaction between the filters. Therefore, it achieves the state-of-the-art performance in computer vision tasks such as object detection\cite{gao2019res2net} and person re-identification\cite{cao2019person}. Thanks to the performance, it inspires implementing Res2Net for multivariate time series analysis.  However, in Res2Net, the whole previous output feature maps are sent to the next group of filters directly which ignores considering how much information from the previous output should be sent. Or in other words, the model lacks the control on the information interaction between different groups of filters. For multivariate time series analysis, as the feature maps contain information from different variables, ignoring the correlations of the feature maps may lead to lacking the considerations on the inter-relationship of the variables. Thus, effective control on the information interaction process is important and meaningful.
\\In this paper, we propose a backbone CNN based on the gated mechanism and Res2Net called Gated Res2Net (GRes2Net). Compared with Res2Net, we utilize gates in the connection between the previous output feature maps and the next input feature maps in order to consider the correlations between different feature maps. Instead of simply executing addition in the residual-like connections, the gates whose values are calculated based on the original feature maps, the previous output feature maps and the input feature maps decide how much information will be sent through the connection. In this way, GRes2Net can control the process of information interaction and utilize the latent information more effectively for a better performance.
\\The contributions of our paper are in two aspects:
\\First, we propose a novel backbone CNN called GRes2Net for multivariate time series analysis. I.e., it can be used for both classification and forecasting. The proposed backbone achieves effectively considering correlations between the variables and extracting multi-scale temporal features of multivariate time series data. Moreover, the proposed backbone can be easily implemented in the existing CNN models or RNN-CNN cascade models.
\\Second, we conduct experiments on four multivariate time series datasets including two multivariate time series classification datasets and two multivariate time series forecasting datasets to assess the proposed backbone. We compare our model with other alternatives. The results demonstrate that our model outperforms the state-of-the-art methods hence indicating the advantage of our model.
\\The rest of this paper is organized as follows. Section II presents related studies and techniques. Section III introduces the structure of GRes2Net. Section IV gives the details of the four datasets and the corresponding state-of-the-art method, the evaluation standards, the experiment settings and the results of the experiments. Section V gives the conclusion and future work.

\section{Related Works}
\subsection{Traditional Machine Learning for Multivariate Time Series Analysis}
Traditional machine learning methods are popular for multivariate time series analysis. Prevailing approaches such as DTW, KNN and SVM are applicable for both classification and forecasting\cite{kim2003financial,wu2006forecasting,gorecki2014non,marcacini2016combining}. SVM is widely used in data mining tasks\cite{liu2002fuzzy} and has been improved in many aspects\cite{sun2018fast,cauwenberghs2001incremental}. Models based on the combination of KNN and DTW suggests strong performances\cite{oehmcke2016knn,hsu2011knn}. Autoregressive Moving Average (ARMA) is a typical method for exploring the rational spectrum of stationary random processes and has been widely used for analyzing stationary time series\cite{brockwell2016introduction}. Ensemble learning methods such as Random Forest (RF) and Gradient Boosting Machine (GBM) aggregate multiple classifiers into one model and generally result in better performance than single classifier. Related methods are commonly used in time series analysis\cite{kane2014comparison} and data science competition\cite{taieb2014gradient}.

\subsection{Gated Mechanism}
Gated mechanism is commonly used in RNN. Because the original RNN suffers for gradient vanishing or explosion. Besides, it is likely to ignore long-term dependencies\cite{chung2014empirical,jozefowicz2015empirical}. Thus, in order to address these problems, gated mechanism is applied. Long Short-Term Memory (LSTM)\cite{hochreiter1997long} and Gated Recurrent Unit (GRU)\cite{cho2014learning} are two well-known variants of RNN based on the thought of gated mechanism. Taking LSTM as an example, specifically, the computation of LSTM is defined by the equations \eqref{1}-\eqref{6}:

\begin{equation}
\label{1}
g^{(k)}=\tanh \left(W^{g x} x_{k}+W^{g h} h_{(k-1)}+b^{g}\right)
\end{equation}
\begin{equation}
\label{2}
i^{(k)}=\sigma\left(W^{i x} x_{k}+W^{i h} h_{(k-1)}+b^{i}\right)
\end{equation}
\begin{equation}
\label{3}
f^{(k)}=\sigma\left(W^{f x} x_{k}+W^{f h} h_{(k-1)}+b^{f}\right)
\end{equation}
\begin{equation}
\label{4}
o^{(k)}=\sigma\left(W^{o x} x_{k}+W^{o h} h_{(k-1)}+b^{o}\right)
\end{equation}
\begin{equation}
\label{5}
s^{(k)}=g^{(k)} \odot i^{(k)}+s^{(k-1)} \odot f^{(k)}
\end{equation}
\begin{equation}
\label{6}
h_{(k)}=\tanh \left(s^{(k)}\right) \odot o^{(k)}
\end{equation}
\\where \emph{W$^{gx}$}, \emph{W$^{gh}$}, \emph{W$^{ix}$}, \emph{W$^{ih}$}, \emph{W$^{fx}$}, \emph{W$^{fh}$}, \emph{W$^{ox}$}, \emph{W$^{oh}$},  \emph{b$^g$}, \emph{b$^i$}, \emph{b$^f$} and \emph{b$^o$} are parameters to be learnt. \emph{i$^{(k)}$}, \emph{o$^{(k)}$}  and \emph{f$^{(k)}$} are three gates which determine how much the previous information will influence the current output and the current hidden state. Concretely, \emph{f$^{(k)}$} determines how much information from the previous hidden states is going to be abandoned. \emph{i$^{(k)}$}  determines how much information will be sent into the current hidden states and {o$^{(k)}$} determines the output of a single LSTM cell. Gated mechanism is helpful in two aspects: First, it is able to avoid gradient vanishing. Second, it allows the model choose whether the previous information is supposed to be memorized or abandoned hence it allows the recurrent layer to capture long-term dependencies more easily. The benefits of the gated mechanism inspires implementing it in other networks.
\subsection{Res2Net}
Res2Net is a recently proposed backbone CNN. To achieve multi-scale available receptive fields, the filters with \emph{n} channels are replaced with \emph{s} groups of filters and each group of filters are with \emph{w} channels. (To avoiding information loss, generally \emph{n}=\emph{s}$\times$\emph{w}.) Groups of filters are connected in a hierarchical residual-like style. The channels of input feature maps are expanded by convolutional layers and then divided into several groups. A group of filters firstly extracts features from a group of input feature maps. The output feature maps are then sent to the next group of filters along with another group of input feature maps. This process repeats several times until all input feature maps are processed. 
\\Specifically, in a single Res2Net module, channel expansion is firstly executed by implementing convolutional layer. Then, the original feature maps are obtained, denoted by \textbf{X}. After that, \textbf{X} is evenly divided into several groups which are denoted by \textbf{x$_i$}, where \emph{i} $\in$ \emph{\{1,2,3….s\}}. Each group is a feature map subset which has the same spatial or temporal size  and \emph{1/s} of channels compared with \textbf{X}. The convolution is denoted by \textbf{K$_i$()}, \textbf{y$_i$} is the output of \textbf{K$_i$()}. Then, the \textbf{y$_i$}  can be written in equation \eqref{7}:
\begin{equation}
\label{7}
\mathbf{y}_{i}=\left\{\begin{array}{ll}
{\mathbf{x}_{i}} & {i=1} \\
{\mathbf{K}_{i}\left(\mathbf{x}_{i}\right)} & {i=2} \\
{\mathbf{K}_{i}\left(\mathbf{x}_{i}+\mathbf{y}_{i-1}\right)} & {2<i \leqslant s.}
\end{array}\right.
\end{equation}
\\All the outputs are concatenated and then fed into convolutional layers for channel compression and information fusion. In this way, Res2Net achieves multi-scale receptive fields thus allowing multi-scale feature representations. The architectures based on Res2Net backbone have achieved the state-of-the-art performance in several computer vision tasks\cite{gao2019res2net}. However, the original structure of Res2Net suffers for controlling the information flow between groups. Hence, we considering implementing gated mechanism in Res2Net to fix that problem.
\section{Gated Res2Net}
The structure shown in Fig.~\ref{figure1}(a) is the original Res2Net module while the structure shown in Fig.~\ref{figure1}(b) is the proposed GRes2Net module. Fig.~\ref{figure1} shows the difference between Res2Net and GRes2Net. Code is available at: \url{ https://github.com/ChaoYang93/GraduatePaper/blob/master/GRes2Net.py}.
\\In our GRes2Net, instead of sending the whole previous output feature maps to the next group of filters along with the another group of input feature maps, gates are utilized to determine how much information should be sent. We use the same denotations in the previous section. Moreover, \textbf{g$_i$} denotes the gate, then the \textbf{y$_i$} of GRes2Net can be presented as equation \eqref{8}:
\begin{equation}
\label{8}
\mathbf{y}_{i}=\left\{\begin{array}{ll}
{\mathbf{x}_{i}} & {i=1} \\
{\mathbf{K}_{i}\left(\mathbf{x}_{i}\right)} & {i=2} \\
{\mathbf{K}_{i}\left(\mathbf{x}_{i}+\mathbf{g}_{i}\cdot\mathbf{y}_{i-1}\right)} & {2<i \leqslant s}
\end{array}\right.
\end{equation}
where for the \textbf{x$_i$}, \textbf{g$_i$} is calculated as equation \eqref{9}:
\begin{equation}
\label{9}
\mathbf{g}_{i}=\tanh \left(a\left(\operatorname{concat}\left(a(\mathbf{X}), a(\mathbf{y}_{i-1}), a\left(\mathbf{x}_{i}\right)\right)\right)\right).
\end{equation}
Generally, \emph{a} can be fully connected layers or convolutional layers. The value of the \textbf{g$_i$} is calculated considering the original feature maps \textbf{X}, the next input feature maps \textbf{x$_i$} and the previous output feature maps \textbf{y$_{i-1}$}. An illustration for calculation can be seen in Fig.~\ref{figure2}.
\\The following computations are the same as Res2Net. All the outputs are concatenated and go through a convolutional layer for channel compression and information fusion. Convolutional layer whose kernel size equals to 1 is used for channel expansion and compression which is beneficial for information fusion without too many extra parameters.
\\For multivariate time series analysis, GRes2Net module uses one-dimension convolutional layers. Each group of the feature maps contains temporal information from different variables. Through the connection between the groups of filters, multi-scale receptive fields are achieved which is beneficial for both global and local temporal features extraction. Besides, based on the gated mechanism, the correlations of different feature maps are also considered in a more effective way compared with Res2Net. The model is able to control the information interaction process which is helpful for better utilizing the correlations between different feature maps. Thus, GRes2Net is capable for multivariate time series analysis. 
\section{Experiments}
\subsection{Datasets Description and the Corresponding State-of-the-art Method}\label{AA}
\subsubsection{EGG Dataset}
EGG dataset is a multivariate time series classification dataset which consists of 64 attributes. The maximum length of a sequence in the dataset is 256. The dataset can be downloaded from\cite{ucl}. It has predefined training and validation sets. The training set contains 600 sequences while the validation set contains the same number of sequences. LSTM-FCN, MLSTM-FCN, ALSTM-FCN and MALSTM-FCN\cite{karim2019multivariate} have been implemented on the dataset for evaluation. Among them, MALSTM-FCN achieves the state-of-the-art performance based on the accuracy.
\\In order to make sure the comparison is fair and reasonable, we use the same data preprocessing method as\cite{karim2019multivariate} and use accuracy to evaluate the performance of our model. We use the same training set to train our model and use the same validation set for evaluation as\cite{karim2019multivariate}.
\subsubsection{Occupancy Detection Dataset}
Occupancy detection dataset is a multivariate time series classification dataset. This dataset is developed by\cite{candanedo2016accurate} and can be obtained from\cite{ucl}. The dataset contains 5 attributes and the maximum length of the sequence in the dataset is 3758. The aim is to detect whether the office is occupied according to the temperature, humidity, light, CO2 and humidity ratio. The dataset has predefined training and validation sets. LSTM-FCN, MLSTM-FCN, ALSTM-FCN and MALSTM-FCN\cite{karim2019multivariate} have been implemented on the dataset and the state-of-the-art performance based on accuracy is achieved by MLSTM-FCN. We use the same data preprocessing method as\cite{karim2019multivariate} and use accuracy as the criterion. Moreover, we use the same training set to train our model and use the same validation set for evaluation as\cite{karim2019multivariate}.
\subsubsection{Appliances Energy Prediction Dataset}
Appliances energy prediction dataset contains experimental data used to create forecasting models of appliances energy use in a low energy building. The dataset contains 19735 measurements. It is proposed by\cite{candanedo2017data} and can be obtained from\cite{ucl}. The dataset contains 29 attributes including temperature, humidity, weather, etc. In\cite{candanedo2017data}, GBM makes the most accurate prediction. The recent state-of-the-art performance is achieved by Multi-Layer Perceptron (MLP) proposed in\cite{chammas2019efficient}. Root Mean Squared Error (RMSE), Mean Absolute Error (MAE), Mean Absolute Percentage Error (MAPE) and R Squared(R$^2$) are used for evaluation. The details of the criteria are shown in equations \eqref{10}-\eqref{13}:
\begin{equation}
\label{10}
RMSE=\sqrt[2]{\frac{1}{M} \sum_{1}^{M}\left(y_{i}-\hat{y}_{i}\right)^{2}}
\end{equation}
\begin{equation}
\label{11}
MAE=\frac{1}{M} \sum_{1}^{M}\left|y_{i}-\hat{y}_{i}\right|
\end{equation}
\begin{equation}
\label{12}
MAPE=\frac{100}{M} \sum_{1}^{M}\left|\frac{y_{i}-\hat{y}_{i}}{y_{i}}\right|
\end{equation}
\begin{equation}
\label{13}
R^{2}=1-\frac{\sum_{1}^{M}\left(y_{i}-\hat{y}_{i}\right)^{2}}{\sum_{1}^{M}\left(y_{i}-\bar{y}_{i}\right)^{2}}
\end{equation}
where \emph{M} denotes the total number of the samples, $y_{i}$ denotes the true value, $\bar{y}_{i}$ denotes the average of the true values and $\hat{y}_{i}$ denotes the output of the model.
\\We use the same data preprocessing method and evaluation criteria as \cite{chammas2019efficient} for reasonable comparison.
\subsubsection{Individual Household Electric Power Consumption Dataset}
Individual household electric power consumption dataset contains measurements of electric power consumption in one household over a period of almost 4 years. This dataset can be obtained from\cite{ucl}. It contains 2075259 measurements gathered in a house located in Sceaux (7km of Paris, France) between December 2006 and November 2010 (47 months). The state-of-the-art performance is achieved by ARMA proposed in\cite{chujai2013time} using RMSE for evaluation. We build our model to predict daily series contains 42 measurements. The result based on ARMA with the same data preprocessing method is given in\cite{chujai2013time}. Moreover, we use RMSE, MAE, MAPE and R$^2$ for evaluation. It is worth mentioning that because the specific MAE, MAPE and R$^2$ are not given in \cite{chujai2013time}, we use N/A instead.
\subsection{Experiment Setting}
In our experiment, we use GRes2Net as the backbone CNN for feature extraction. \emph{a} is convolutional layer in the experiment. The output feature maps go through a global average pooling layer and fully connected layers to learn the mapping between the feature maps and the output. For classification tasks, we use cross-entropy as the lost function while for forecasting tasks we use Mean Squared Error (MSE). We train our model by an Adam optimizer with the initial learning rate 0.001 for totally 500 training epochs. Adam is a commonly used optimizer and generally has the best performance, so we did not try others. After each 100 epochs, the learning rate is adjusted to one tenth of the previous value. The batch size is 32 in the two classification tasks and the batch size is 64 in the two forecasting tasks. Dropout is used to avoid overfitting. The random invalid possibility of the neurons is 0.5. The early stopping is performed based on the validation error. I.e., we store the model that has the best performance on the validation set. Then, we repeat validating on the validation set for 5 times and record the results. To avoid drastic fluctuation of the performance, we calculate the average performance as the final result.
\\Besides, we implement deep LSTM and Res2Net on the four datasets for comparison. We use the same settings to train LSTM and Res2Net. The model based on Res2Net contains the same number of modules as the model based on GRes2Net for fair comparison. We construct a bidirectional LSTM contains 8 layers. Dropout is also used and the random invalid possibility is the same. Early stopping and 5 times validating are also executed.
\\We compare our model with the aforementioned state-of-the-art models, deep LSTM, Res2Net and other related models that have been implemented in the previous studies. We implement LSTM, Res2Net and GRes2Net with PyTorch, a famous and popular deep learning framework based on Python. All models are trained with GPU GTX 1060.
\subsection{Experiment Results}
Table~\ref{table1} shows the classification results. As we can see, GRes2Net outperforms the alternatives on the two multivariate time series classification datasets. It is worth noticing that the previous state-of-the-art models are based on the variant of RNN-CNN cascade model. But original Res2Net and GRes2Net both have a better performance than the previous state-of-the-art models which indicates the advantages of the hierarchical residual-like connections for multivariate time series classification. Moreover, it is likely that replacing the CNN that used in the cascade model with GRes2Net may improve the performance.
\\ Table~\ref{table2} and Table~\ref{table3} demonstrate that GRes2Net has the best performance on the two multivariate time series forecasting datasets according to the evaluation criteria. Several traditional machine learning algorithms are implemented for forecasting. It is obvious that GRes2Net has a significant improvement compared with the previous state-of-the-art methods. Surprisingly, LSTM is supposed to have a better performance than MLP on appliances energy prediction dataset while the result is opposite. It is possible that some tricks have been used for constructing MLP in the previous work.
\\According to the results of the experiments, Res2Net outperforms the state-of-the-art models on all four datasets. As mentioned in the previous section, in a Res2Net module, the previous output features maps are sent to the next group of filters without any restriction. However, for multivariate time series analysis, the drawback of the Res2Net is that, if simply sending all the previous output feature maps to the next group of filters, it is incapable for the model to control how much information is supposed to be sent. Hence, it is likely that the some connections have no or even negative influence for multivariate time series analysis. 
\\To address that problem, gated mechanism is used. The gates play an important role in GRes2Net as it allows the model to control the process of information interactions between the different groups of filters. Concretely, the connections between the groups of filters are influenced by the correlations of the feature maps. Thanks to that, the model is able to determine how much information will be sent for achieving a better performance. In this way, for multivariate time series analysis, not only a multi-scale temporal features are extracted, but the correlations between the feature maps are more effectively considered. It is conducive for accurate multivariate time series analysis. This can be easily seen according to the results of the experiments as the performance of GRes2Net is the best on all four datasets. Besides, as the sizes of the four datasets are quite different and all the experimental results are satisfying, it is convincing to say that GRes2Net is able to perform well among different datasets. To conclude, GRes2Net demonstrates the superiority over Res2Net and other state-of-the-art models for multivariate time series analysis.
\begin{table}[]
	\centering
	\renewcommand{\arraystretch}{1.2}
	\caption{Performance Comparison for Classification Tasks}
	\label{table1}
	\resizebox{0.48\textwidth}{!}{%
		\begin{tabular}{|c|c|c|}
			\hline
			Method & EGG (Accuracy) & Occupancy Detection (Accuracy) \\ \hline
			LSTM & 71.60 & 57.99 \\ \hline
			Res2Net & 91.50 & 81.25 \\ \hline
			LSTM-FCN & 90.67 & 71.05 \\ \hline
			MLSTM-FCN & 91.00 & 76.31 \\ \hline
			ALSTM-FCN & 90.67 & 71.05 \\ \hline
			MALSTM-FCN & 91.33 & 72.37 \\ \hline
			\textbf{GRes2Net} & \textbf{92.76} & \textbf{83.33} \\ \hline
		\end{tabular}%
	}
\end{table}
\begin{table}[]
	\centering
	\renewcommand{\arraystretch}{1.2}
	\caption{Performance Comparison Based on Appliances Energy Prediction Dataset}
	\label{table2}
	\resizebox{0.37\textwidth}{!}{%
		\begin{tabular}{|c|c|c|c|c|}
			\hline
			Method & RMSE & MAE & MAPE & R$^2$ \\ \hline
			GBM & 66.65 & 35.22 & 38.29 & 0.57 \\ \hline
			RF & 68.48 & 31.85 & 31.39 & 0.54 \\ \hline
			SVM & 70.74 & 31.36 & 29.76 & 0.52 \\ \hline
			LSTM & 74.41 & 39.21 & 41.08 & 0.24 \\ \hline
			MLP & 59.84 & 27.28 & 27.09 & 0.64 \\ \hline
			Res2Net & 13.98 & 7.64 & 10.47 & 0.97 \\ \hline
			\textbf{GRes2Net} & \textbf{12.84} & \textbf{6.99} & \textbf{9.74} & \textbf{0.98} \\ \hline
		\end{tabular}%
	}
\end{table}
\begin{table}[]
	\centering
	\renewcommand{\arraystretch}{1.2}
	\caption{Performance Comparison Based on Individual Household Electric Power Consumption Dataset}
	\label{table3}
	\resizebox{0.38\textwidth}{!}{%
		\begin{tabular}{|c|c|c|c|c|}
			\hline
			Method & RMSE & MAE & MAPE & R$^2$ \\ \hline
			LSTM & 0.61 & 0.39 & 41.49 & 0.72 \\ \hline
			Res2Net & 0.20 & 0.13 & 13.13 & 0.97 \\ \hline
			ARMA & 0.36 & N/A & N/A & N/A \\ \hline
			\textbf{GRes2Net} & \textbf{0.19} & \textbf{0.12} & \textbf{12.88} & \textbf{0.98} \\ \hline
		\end{tabular}%
	}
\end{table}
\section{Conclusion and Future Work}
In this paper, we propose GRes2Net, a backbone CNN for multivariate time series analysis based on gated mechanism and Res2Net. In our model, instead of sending the whole previous output feature maps to the next input, gates are used to control this process. The value of a single gate is calculated based on the undivided feature maps, the previous output feature maps and the next input feature maps. Then the gate determines how much information to be sent. In this way, our model is able to consider the correlations between the feature maps more effectively compared with Res2Net. Moreover, it is worth mentioning that GRes2Net can be implemented in the existing CNN models or RNN-CNN cascade models for multivariate time series analysis without extra effort. Experiments results indicate that our model achieves more accurate results for both classification and forecasting than the state-of-the-art models hence demonstrating the superiority.
\\In the future, we will further explore the design of the gates. Concretely, we are going to consider calculating the value of the gates based on different variables. Besides, the original feature maps are divided evenly. This division method is likely to be improved. Thus, we are going to explore a more effective way such as splitting by learning. 

\section*{Acknowledgment}
This research was supported by National Natural Science Foundation of China under Grant 61827810.

\bibliographystyle{IEEEtran} 
\bibliography{IEEEabrv,test}

\begin{thebibliography}{10}
\providecommand{\url}[1]{#1}
\csname url@samestyle\endcsname
\providecommand{\newblock}{\relax}
\providecommand{\bibinfo}[2]{#2}
\providecommand{\BIBentrySTDinterwordspacing}{\spaceskip=0pt\relax}
\providecommand{\BIBentryALTinterwordstretchfactor}{4}
\providecommand{\BIBentryALTinterwordspacing}{\spaceskip=\fontdimen2\font plus
\BIBentryALTinterwordstretchfactor\fontdimen3\font minus
  \fontdimen4\font\relax}
\providecommand{\BIBforeignlanguage}[2]{{%
\expandafter\ifx\csname l@#1\endcsname\relax
\typeout{** WARNING: IEEEtran.bst: No hyphenation pattern has been}%
\typeout{** loaded for the language `#1'. Using the pattern for}%
\typeout{** the default language instead.}%
\else
\language=\csname l@#1\endcsname
\fi
#2}}
\providecommand{\BIBdecl}{\relax}
\BIBdecl

\bibitem{keogh2003need}
E.~Keogh and S.~Kasetty, ``On the need for time series data mining benchmarks:
  a survey and empirical demonstration,'' \emph{Data Mining and knowledge
  discovery}, vol.~7, no.~4, pp. 349--371, 2003.

\bibitem{anguita2012human}
D.~Anguita, A.~Ghio, L.~Oneto, X.~Parra, and J.~L. Reyes-Ortiz, ``Human
  activity recognition on smartphones using a multiclass hardware-friendly
  support vector machine,'' in \emph{International workshop on ambient assisted
  living}.\hskip 1em plus 0.5em minus 0.4em\relax Springer, 2012, pp. 216--223.

\bibitem{liu2018time}
C.-L. Liu, W.-H. Hsaio, and Y.-C. Tu, ``Time series classification with
  multivariate convolutional neural network,'' \emph{IEEE Transactions on
  Industrial Electronics}, vol.~66, no.~6, pp. 4788--4797, 2018.

\bibitem{masethe2014prediction}
H.~D. Masethe and M.~A. Masethe, ``Prediction of heart disease using
  classification algorithms,'' in \emph{Proceedings of the world Congress on
  Engineering and computer Science}, vol.~2, 2014, pp. 22--24.

\bibitem{kolehmainen2001neural}
M.~Kolehmainen, H.~Martikainen, and J.~Ruuskanen, ``Neural networks and
  periodic components used in air quality forecasting,'' \emph{Atmospheric
  Environment}, vol.~35, no.~5, pp. 815--825, 2001.

\bibitem{doucoure2016time}
B.~Doucoure, K.~Agbossou, and A.~Cardenas, ``Time series prediction using
  artificial wavelet neural network and multi-resolution analysis: Application
  to wind speed data,'' \emph{Renewable Energy}, vol.~92, pp. 202--211, 2016.

\bibitem{le2019improving}
T.~Le, M.~T. Vo, B.~Vo, E.~Hwang, S.~Rho, and S.~W. Baik, ``Improving electric
  energy consumption prediction using cnn and bi-lstm,'' \emph{Applied
  Sciences}, vol.~9, no.~20, p. 4237, 2019.

\bibitem{freeman2018forecasting}
B.~S. Freeman, G.~Taylor, B.~Gharabaghi, and J.~Th{\'e}, ``Forecasting air
  quality time series using deep learning,'' \emph{Journal of the Air \& Waste
  Management Association}, vol.~68, no.~8, pp. 866--886, 2018.

\bibitem{che2018recurrent}
Z.~Che, S.~Purushotham, K.~Cho, D.~Sontag, and Y.~Liu, ``Recurrent neural
  networks for multivariate time series with missing values,'' \emph{Scientific
  reports}, vol.~8, no.~1, p. 6085, 2018.

\bibitem{malhotra2017timenet}
P.~Malhotra, V.~TV, L.~Vig, P.~Agarwal, and G.~Shroff, ``Timenet: Pre-trained
  deep recurrent neural network for time series classification,'' \emph{arXiv
  preprint arXiv:1706.08838}, 2017.

\bibitem{gehring2017convolutional}
J.~Gehring, M.~Auli, D.~Grangier, D.~Yarats, and Y.~N. Dauphin, ``Convolutional
  sequence to sequence learning,'' in \emph{Proceedings of the 34th
  International Conference on Machine Learning-Volume 70}.\hskip 1em plus 0.5em
  minus 0.4em\relax JMLR. org, 2017, pp. 1243--1252.

\bibitem{zhao2017convolutional}
B.~Zhao, H.~Lu, S.~Chen, J.~Liu, and D.~Wu, ``Convolutional neural networks for
  time series classification,'' \emph{Journal of Systems Engineering and
  Electronics}, vol.~28, no.~1, pp. 162--169, 2017.

\bibitem{simonyan2014very}
K.~Simonyan and A.~Zisserman, ``Very deep convolutional networks for
  large-scale image recognition,'' \emph{arXiv preprint arXiv:1409.1556}, 2014.

\bibitem{szegedy2015going}
C.~Szegedy, W.~Liu, Y.~Jia, P.~Sermanet, S.~Reed, D.~Anguelov, D.~Erhan,
  V.~Vanhoucke, and A.~Rabinovich, ``Going deeper with convolutions,'' in
  \emph{Proceedings of the IEEE conference on computer vision and pattern
  recognition}, 2015, pp. 1--9.

\bibitem{he2016deep}
K.~He, X.~Zhang, S.~Ren, and J.~Sun, ``Deep residual learning for image
  recognition,'' in \emph{Proceedings of the IEEE conference on computer vision
  and pattern recognition}, 2016, pp. 770--778.

\bibitem{long2015fully}
J.~Long, E.~Shelhamer, and T.~Darrell, ``Fully convolutional networks for
  semantic segmentation,'' in \emph{Proceedings of the IEEE conference on
  computer vision and pattern recognition}, 2015, pp. 3431--3440.

\bibitem{fawaz2019deep}
H.~I. Fawaz, G.~Forestier, J.~Weber, L.~Idoumghar, and P.-A. Muller, ``Deep
  learning for time series classification: a review,'' \emph{Data Mining and
  Knowledge Discovery}, vol.~33, no.~4, pp. 917--963, 2019.

\bibitem{gao2019res2net}
S.-H. Gao, M.-M. Cheng, K.~Zhao, X.-Y. Zhang, M.-H. Yang, and P.~Torr,
  ``Res2net: A new multi-scale backbone architecture,'' \emph{arXiv preprint
  arXiv:1904.01169}, 2019.

\bibitem{cao2019person}
Z.~Cao and H.~J. Lee, ``Person re-identification based on res2net network,''
  \emph{arXiv preprint arXiv:1910.04061}, 2019.

\bibitem{kim2003financial}
K.-j. Kim, ``Financial time series forecasting using support vector machines,''
  \emph{Neurocomputing}, vol.~55, no. 1-2, pp. 307--319, 2003.

\bibitem{wu2006forecasting}
W.~Wu, J.~Zhou, L.~Mo, and C.~Zhu, ``Forecasting electricity market price
  spikes based on bayesian expert with support vector machines,'' in
  \emph{International Conference on Advanced Data Mining and
  Applications}.\hskip 1em plus 0.5em minus 0.4em\relax Springer, 2006, pp.
  205--212.

\bibitem{gorecki2014non}
T.~G{\'o}recki and M.~{\L}uczak, ``Non-isometric transforms in time series
  classification using dtw,'' \emph{Knowledge-based systems}, vol.~61, pp.
  98--108, 2014.

\bibitem{marcacini2016combining}
R.~M. Marcacini, J.~C. Carnevali, and J.~Domingos, ``On combining websensors
  and dtw distance for knn time series forecasting,'' in \emph{2016 23rd
  International Conference on Pattern Recognition (ICPR)}.\hskip 1em plus 0.5em
  minus 0.4em\relax IEEE, 2016, pp. 2521--2525.

\bibitem{liu2002fuzzy}
Y.~Liu and H.~Huang, ``Fuzzy support vector machines for pattern recognition
  and data mining,'' \emph{International journal of fuzzy systems}, vol.~4,
  no.~3, pp. 826--835, 2002.

\bibitem{sun2018fast}
Z.~Sun, K.~Hu, T.~Hu, J.~Liu, and K.~Zhu, ``Fast multi-label low-rank
  linearized svm classification algorithm based on approximate extreme
  points,'' \emph{IEEE Access}, vol.~6, pp. 42\,319--42\,326, 2018.

\bibitem{cauwenberghs2001incremental}
G.~Cauwenberghs and T.~Poggio, ``Incremental and decremental support vector
  machine learning,'' in \emph{Advances in neural information processing
  systems}, 2001, pp. 409--415.

\bibitem{oehmcke2016knn}
S.~Oehmcke, O.~Zielinski, and O.~Kramer, ``knn ensembles with penalized dtw for
  multivariate time series imputation,'' in \emph{2016 International Joint
  Conference on Neural Networks (IJCNN)}.\hskip 1em plus 0.5em minus
  0.4em\relax IEEE, 2016, pp. 2774--2781.

\bibitem{hsu2011knn}
H.-H. Hsu, A.~C. Yang, and M.-D. Lu, ``Knn-dtw based missing value imputation
  for microarray time series data,'' \emph{Journal of computers}, vol.~6,
  no.~3, pp. 418--425, 2011.

\bibitem{kane2014comparison}
M.~J. Kane, N.~Price, M.~Scotch, and P.~Rabinowitz, ``Comparison of arima and
  random forest time series models for prediction of avian influenza h5n1
  outbreaks,'' \emph{BMC bioinformatics}, vol.~15, no.~1, p. 276, 2014.

\bibitem{taieb2014gradient}
S.~B. Taieb and R.~J. Hyndman, ``A gradient boosting approach to the kaggle
  load forecasting competition,'' \emph{International journal of forecasting},
  vol.~30, no.~2, pp. 382--394, 2014.

\bibitem{brockwell2016introduction}
P.~J. Brockwell and R.~A. Davis, \emph{Introduction to time series and
  forecasting}.\hskip 1em plus 0.5em minus 0.4em\relax springer, 2016.

\bibitem{chung2014empirical}
J.~Chung, C.~Gulcehre, K.~Cho, and Y.~Bengio, ``Empirical evaluation of gated
  recurrent neural networks on sequence modeling,'' \emph{arXiv preprint
  arXiv:1412.3555}, 2014.

\bibitem{jozefowicz2015empirical}
R.~Jozefowicz, W.~Zaremba, and I.~Sutskever, ``An empirical exploration of
  recurrent network architectures,'' in \emph{International Conference on
  Machine Learning}, 2015, pp. 2342--2350.

\bibitem{hochreiter1997long}
S.~Hochreiter and J.~Schmidhuber, ``Long short-term memory,'' \emph{Neural
  computation}, vol.~9, no.~8, pp. 1735--1780, 1997.

\bibitem{cho2014learning}
K.~Cho, B.~Van~Merri{\"e}nboer, C.~Gulcehre, D.~Bahdanau, F.~Bougares,
  H.~Schwenk, and Y.~Bengio, ``Learning phrase representations using rnn
  encoder-decoder for statistical machine translation,'' \emph{arXiv preprint
  arXiv:1406.1078}, 2014.

\bibitem{ucl}
M.~Lichman, ``Uci machine learning repository,''
  \url{http://archive.ics.uci.edu/ml}, accessed 2013.

\bibitem{karim2019multivariate}
F.~Karim, S.~Majumdar, H.~Darabi, and S.~Harford, ``Multivariate lstm-fcns for
  time series classification,'' \emph{Neural Networks}, vol. 116, pp. 237--245,
  2019.

\bibitem{candanedo2016accurate}
L.~M. Candanedo and V.~Feldheim, ``Accurate occupancy detection of an office
  room from light, temperature, humidity and co2 measurements using statistical
  learning models,'' \emph{Energy and Buildings}, vol. 112, pp. 28--39, 2016.

\bibitem{candanedo2017data}
L.~M. Candanedo, V.~Feldheim, and D.~Deramaix, ``Data driven prediction models
  of energy use of appliances in a low-energy house,'' \emph{Energy and
  buildings}, vol. 140, pp. 81--97, 2017.

\bibitem{chammas2019efficient}
M.~Chammas, A.~Makhoul, and J.~Demerjian, ``An efficient data model for energy
  prediction using wireless sensors,'' \emph{Computers \& Electrical
  Engineering}, vol.~76, pp. 249--257, 2019.

\bibitem{chujai2013time}
P.~Chujai, N.~Kerdprasop, and K.~Kerdprasop, ``Time series analysis of
  household electric consumption with arima and arma models,'' in
  \emph{Proceedings of the International MultiConference of Engineers and
  Computer Scientists}, vol.~1, 2013, pp. 295--300.

\end{thebibliography}

\end{document}